\documentclass[runningheads]{llncs}

\usepackage[T1]{fontenc} 
\usepackage{graphicx}    
\usepackage{amsmath}
\usepackage{amssymb}
\usepackage{amsfonts}
\usepackage{multirow}
\usepackage{array}
\usepackage{booktabs}
\usepackage{hyperref}    
\usepackage{color}
\hypersetup{
    colorlinks=true,
    linkcolor=blue,
    filecolor=magenta,      
    urlcolor=blue,
    citecolor=blue,
}
\usepackage{pifont}
\usepackage{url}
\usepackage{adjustbox}
\usepackage{pdflscape}
\usepackage{anyfontsize}
\newcommand{\repeatthanks}{\textsuperscript{\thefootnote}}
\usepackage{soul}
\usepackage[moderate,tracking=normal,paragraphs=normal]{savetrees} 

\begin{document}

\title{CAGN-GAT Fusion: A Hybrid Contrastive Attentive Graph Neural Network for Network Intrusion Detection}



\author{Md Abrar Jahin\inst{1,2}\thanks{Authors contributed equally}
\and Shahriar Soudeep\inst{3}\repeatthanks
\and Fahmid Al Farid\inst{4}
\and M. F. Mridha\inst{3}
\and Raihan Kabir\inst{5}
\and Md Rashedul Islam\inst{6}
\and Hezerul Abdul Karim\inst{4}
}

\authorrunning{Md Abrar Jahin et al.}
\institute{Thomas Lord Department of Computer Science, University of Southern California, Los Angeles, CA 90007, USA\\
\email{abrar.jahin.2652@gmail.com, jahin@usc.edu}
\and Physics and Biology Unit, Okinawa Institute of Science and Technology Graduate University (OIST), Okinawa 904-0412, Japan\\
\email{mdabrar.jahin@oist.jp}
\and Department of Computer Science, American International University-Bangladesh, Dhaka 1229, Bangladesh\\
\email{20-43823-2@student.aiub.edu, firoz.mridha@aiub.edu}
\and Centre for Image and Vision Computing (CIVC), COE for Artificial Intelligence, Faculty of Artificial Intelligence and Engineering (FAIE), Multimedia University, Cyberjaya 63100, Selangor, Malaysia\\
\email{fahmid.farid@mmu.edu.my, hezerul@mmu.edu.my}
\and Department of Computer Science and Engineering, University of Aizu, Aizuwakamatsu, Fukushima, Japan\\
\email{raihan.kabir.cse@gmail.com}
\and Department of Computer Science and Engineering, University of Asia Pacific, Dhaka 1205, Bangladesh\\
\email{rashed.cse@gmail.com}
}

\maketitle

\begin{abstract}
Cybersecurity threats are growing, making network intrusion detection essential. Traditional machine learning models remain effective in resource-limited environments due to their efficiency, requiring fewer parameters and less computational time. However, handling short and highly imbalanced datasets remains challenging. In this study, we propose the fusion of \ul{C}ontrastive \ul{A}ttentive \ul{G}raph \ul{N}etwork and \ul{G}raph \ul{At}tention Network (CAGN-GAT Fusion), and benchmark it against 15 other models, including both Graph Neural Networks (GNNs) and traditional ML models. Our evaluation is conducted on four benchmark datasets (KDD-CUP-1999, NSL-KDD, UNSW-NB15, and CICIDS2017) using a short and proportionally imbalanced dataset with a constant size of 5000 samples to ensure fairness in comparison. Results show that CAGN-GAT Fusion demonstrates stable and competitive accuracy, recall, and F1-score, even though it does not achieve the highest performance in every dataset. Our analysis also highlights the impact of adaptive graph construction techniques, including small changes in connections (edge perturbation) and selective hiding of features (feature masking), improving detection performance. The findings confirm that GNNs, particularly CAGN-GAT Fusion, are robust and computationally efficient, making them well-suited for resource-constrained environments. Future work will explore GraphSAGE layers and multiview graph construction techniques to further improve adaptability and detection accuracy.

\end{abstract}

\section{Introduction}

The rapid evolution of network technologies and the exponential growth of internet-connected devices have significantly increased the complexity and volume of network traffic. As a result, modern networks are increasingly vulnerable to sophisticated cyber threats, including denial-of-service (DoS) attacks, data exfiltration, and advanced persistent threats. Traditional security mechanisms, such as firewalls and signature-based intrusion detection systems (IDS), often do not identify new and evolving attack patterns. To address these challenges, machine learning (ML)- based IDS have been proposed to enhance detection capabilities by learning patterns from historical data and identifying anomalies in real-time.

Although ML models such as Random Forest (RF), Support Vector Machines (SVM), and XGBoost have demonstrated effectiveness in intrusion detection, they face limitations in capturing the intricate relationships between network entities and attack behaviors \cite{choudhury_comparative_2015}. Recently, Graph Neural Networks (GNNs) have emerged as a promising approach for network intrusion detection due to their ability to model complex network structures and exploit relational dependencies among data points \cite{zhong_survey_2024}. Among various GNN architectures, Graph Convolutional Networks (GCN), Graph Attention Networks (GAT), GraphSAGE, and Graph Isomorphism Networks (GIN) have gained traction in cybersecurity applications \cite{bilot_graph_2023}.

To address these challenges, this study evaluates GNN-based IDS and introduces a novel approach to enhance detection performance. Our findings show that CAGN-GAT Fusion achieves strong accuracy, recall, and F1-score while maintaining computational efficiency. Unlike prior works, this study focuses purely on performance evaluation without integrating Explainable AI (XAI) techniques \cite{zhang_explainable_2022}.

This study introduces CAGN-GAT Fusion, a novel fusion of Contrastive Attentive Graph Network (CAGN) and GAT, demonstrating robust and stable performance in network intrusion detection under a resource-constrained environment. A comprehensive benchmarking effort against 15 existing models, spanning both traditional ML and GNN approaches, highlights the generalizability and competitiveness of the proposed method. Additionally, the study incorporates adaptive graph construction techniques while systematically analyzing the impact of edge perturbations and feature masking on model performance.

The rest of this paper is structured as follows. Section~\ref{sec:related_work} reviews related works on intrusion detection using ML and GNN models. Section~\ref{sec:methodology} details the proposed models and experimental setup. Section~\ref{sec:resultdiscussion} analyzes experimental results. Finally, Section~\ref{sec:conclusion} provides concluding remarks and potential directions for future research.

\section{Related Works}
\label{sec:related_work}

Extensive research has been conducted on network intrusion detection using traditional ML, deep learning (DL), and, more recently, graph-based approaches. Traditional ML models such as SVM, RF, Decision Trees (DT), and XGBoost have gained popularity for their efficiency in classifying network traffic \cite{halimaa_machine_2019}. However, these methods often depend heavily on manual feature engineering and face challenges in generalizing to novel or evolving attack patterns \cite{dong_comparison_2016}. Moreover, their performance tends to degrade on high-dimensional and imbalanced datasets, limiting their practicality in real-time intrusion detection scenarios \cite{pietraszek_data_2005}.

To address these limitations, GNNs have recently gained significant attention for their ability to model the complex structural relationships inherent in network traffic data. Popular architectures such as GCN, GAT, GraphSAGE, and Graph Isomorphism Networks (GIN) have consistently outperformed traditional machine learning models in this domain \cite{ma_comprehensive_2021}. GCNs aggregate information from neighboring nodes through convolutional operations, while GATs introduce attention mechanisms to assign varying importance to different neighbors during message passing \cite{wu_graph_2021}. Further advancements like GraphSAGE and GIN enhance scalability and representation quality through neighbor sampling and refined aggregation strategies, respectively \cite{pujol-perich_unveiling_2022}. However, most existing studies fall short of comprehensively comparing multiple GNN architectures under consistent experimental conditions and often neglect the critical influence of graph construction strategies on model performance \cite{zhou_hierarchical_2021}. 

Moreover, only a limited number of works have systematically benchmarked GNN-based intrusion detection systems against traditional ML models. Recent research has explored self-supervised GNNs for intrusion detection \cite{Xu_2024}, unsupervised graph-based user behavior modeling in social networks \cite{Tran_2022}, and traditional ML techniques for efficient ECG-based atrial fibrillation detection \cite{nguyen_2024}. Many of these studies either focus on a single GNN model or incorporate XAI techniques, which, while valuable for interpretability, can shift focus away from thoroughly assessing detection performance \cite{li_advancing_2024,longa_explaining_2025}. Consequently, the existing literature lacks a clear, comparative understanding of how different GNN architectures perform relative to one another and to classical ML baselines across diverse network intrusion detection scenarios.

Furthermore, several persistent challenges remain in this research area. Scalability and computational efficiency continue to hinder the deployment of both traditional ML and DL models in large-scale, real-time environments \cite{macas_survey_2022}. The adaptability of these models to emerging and evolving cyber threats also remains a concern, as most approaches exhibit limited generalization to unseen attack types \cite{maddireddy_adaptive_2023}. Finally, the overemphasis on interpretability solutions in some studies can obscure the primary objective of maximizing detection performance \cite{lansky_deep_2021}. Motivated by these gaps, our study proposes a novel CAGN-GAT Fusion model, benchmarks it comprehensively against 15 baseline models across multiple datasets, and critically examines the effects of graph construction strategies and perturbation techniques on intrusion detection accuracy. 

By addressing these challenges, our research contributes a rigorous performance-driven analysis of GNN-based intrusion detection models compared to traditional ML techniques. The following section details our proposed methodology and experimental setup.

\section{The Contrastive Attentive Graph Network and Graph Attention Network}
\label{sec:methodology}

This section details the methodology employed in our study, focusing on transforming network intrusion data into graph representations, designing GNN architectures, and the experimental setup. The proposed framework constructs multiple graph structures from network traffic, processes these graphs using advanced GNN models, and evaluates their effectiveness against traditional ML models.

\subsection{Graph Construction Strategies}
A key challenge in intrusion detection is effectively modeling network data. We employ two graph construction strategies, each designed to enhance the structural representation of network traffic data.

\subsubsection{Adaptive Graph Construction}: 
The adaptive graph construction method dynamically creates graph structures based on feature similarity or domain knowledge. Given a feature matrix \( X \in \mathbb{R}^{N \times d} \), where \( N \) represents the number of nodes and \( d \) denotes the feature dimensions, we compute pairwise distances using a selected similarity metric, such as Euclidean or cosine distance. Mathematically, the pairwise Euclidean distance between two nodes \( i \) and \( j \) is computed as:
\begin{equation}
D_{ij} = \|X_i - X_j\|_2
\end{equation}
where \( D_{ij} \) is the computed distance. A binary adjacency matrix \( A \) is then formed by thresholding these distances:
\begin{equation}
A_{ij} = 
\begin{cases} 
1, & \text{if } D_{ij} < \tau \\
0, & \text{otherwise}
\end{cases}
\end{equation}
where \( \tau \) is a user-defined similarity threshold. Additionally, the adjacency matrix is refined using a k-nearest neighbors (KNN) graph, ensuring meaningful connectivity. The final edge index is extracted from \( A \), and the data is returned as a PyTorch Geometric `Data' object containing node features, edge indices, and labels.

\subsubsection{Adaptive Graph with Augmentation}: 
The graph augmentation method introduces controlled perturbations to the graph structure and features to enhance robustness. Edge perturbation involves randomly selecting a fraction of existing edges and duplicating them to introduce structural noise. If the total number of edges is \( |E| \), then the number of perturbed edges is defined as:
\begin{equation}
|E'| = |E| + \lfloor r_e \cdot |E| \rfloor
\end{equation}
where \( r_e \) is the edge perturbation rate. Feature masking is applied to a randomly selected fraction of node features. If the feature matrix has \( d \) dimensions per node, then the number of masked features is:
\begin{equation}
d' = d - \lfloor r_f \cdot d \rfloor
\end{equation}
where \( r_f \) is the feature mask rate, masked features are set to zero. The augmented graph, with its modified edges and features, is then returned as an updated PyTorch Geometric `Data' object.

\subsection{Graph Neural Network Architectures}
In our study, we implement and evaluate several Graph Neural Network (GNN) architectures for network intrusion detection, each designed with specific configurations to balance complexity and performance within our page constraints.

\subsubsection{Graph Convolutional Network (GCN)}:
GCN \cite{zhang_graph_2019} follows a spectral approach to aggregate neighborhood information. The propagation rule for each layer is defined as:
\begin{equation}
H^{(l+1)} = \sigma(\tilde{D}^{-\frac{1}{2}} \tilde{A} \tilde{D}^{-\frac{1}{2}} H^{(l)} W^{(l)})
\end{equation}
where \( \tilde{A} = A + I \) is the adjacency matrix with self-loops, \( \tilde{D} \) is the degree matrix, and \( W^{(l)} \) is the learnable weight matrix. 

\subsubsection{Graph Attention Network (GAT)}:
GAT \cite{velickovic_graph_2018} enhances node feature aggregation using self-attention mechanisms. The attention coefficient between nodes \( i \) and \( j \) is computed as:
\begin{equation}
\alpha_{ij} = \frac{\exp(LeakyReLU(a^T [W h_i || W h_j]))}{\sum_{k \in \mathcal{N}(i)} \exp(LeakyReLU(a^T [W h_i || W h_k]))}
\end{equation}
where \( a \) and \( W \) are trainable parameters.

\subsubsection{Graph Isomorphism Network (GIN)}:
GIN \cite{zhang_graph_2024} follows a message-passing paradigm with a learnable function:
\begin{equation}
h_v^{(k)} = MLP^{(k)} \left((1 + \epsilon) h_v^{(k-1)} + \sum_{u \in \mathcal{N}(v)} h_u^{(k-1)} \right)
\end{equation}
where \( \epsilon \) is a learnable parameter.

\subsubsection{SuperGAT}:
SuperGAT extends the GAT architecture by incorporating self-supervised learning techniques to enhance attention mechanisms. Our implementation consists of three SuperGATConv layers with a configuration similar to the standard GAT model: the first layer has 2 attention heads with 64 hidden units each, followed by single-head layers maintaining the hidden dimension. This setup aims to improve the model's ability to focus on critical connections in the network data.

\subsubsection{GraphSAGE}:
GraphSAGE \cite{hamilton_inductive_2017} uses neighborhood sampling to improve scalability. It computes node embeddings using an aggregation function such as mean, LSTM, or max-pooling:
\begin{equation}
h^{(k)}_v = \sigma(W^{(k)} \cdot AGGREGATE({h^{(k-1)}_u, \forall u \in \mathcal{N}(v)}))
\end{equation}
where \( AGGREGATE \) represents a chosen function.

\subsubsection{Cluster-GCN}:
ClusterGCN \cite{chiang_cluster-gcn_2019} partitions large graphs into clusters and applies GCN within each cluster for efficient training. The adjacency matrix is block-partitioned to allow mini-batch training.

\subsubsection{Auto-Regressive Moving Average (ARMA) GNN}:
ARMA GCN \cite{bianchi_graph_2021} stacks multiple ARMA (Auto-Regressive Moving Average) filters to refine node embeddings:
\begin{equation}
H^{(l+1)} = \sigma(\sum_{k=1}^{K} W_k H^{(l)})
\end{equation}
where \( K \) controls the number of stacked ARMA filters.

\subsubsection{Contrastive Attentive Graph Network (CAGN)}:
CAGN introduces contrastive learning into graph attention networks by pulling similar nodes closer while pushing dissimilar nodes apart. The contrastive loss is formulated as:
\begin{equation}
\mathcal{L}_{contrast} = \sum_{(i,j) \in P} (1 - \cos(h_i, h_j)) + \sum_{(i,j) \in N} \max(0, \cos(h_i, h_j) - \delta)
\end{equation}
where \( \cos(h_i, h_j) \) is the cosine similarity between embeddings \( h_i \) and \( h_j \). The first summation over \( P \) (positive pairs) encourages same-class nodes to have higher similarity. The second summation over \( N \) (negative pairs) applies a margin-based penalty to dissimilar nodes, pushing them apart by a margin \( \delta \). The contrastive framework is implemented via a memory bank that maintains positive and negative sample embeddings, facilitating robust optimization.

\subsubsection{MultiScaleGAT}:
MultiScaleGAT extends GAT by incorporating multi-scale neighborhood information. It utilizes different attention heads at various scales (e.g., local, mid-range, and global neighborhoods) to adaptively learn features at multiple levels:
\begin{equation}
h_v = \sum_{s \in S} \sum_{u \in \mathcal{N}_s(v)} \alpha_{vu}^{(s)} W^{(s)} h_u
\end{equation}
where \( S \) represents different scales of aggregation. The key enhancement in MultiScaleGAT is the introduction of scale-specific attention mechanisms, ensuring that local and global node relationships are effectively captured. Implementation-wise, multiple GAT layers are stacked, each processing information at a distinct scale, and final representations are fused using a weighted sum approach.

\subsubsection{CAGN-GAT Fusion}:
CAGN-GAT Fusion is a hybrid model that integrates CAGN's contrastive learning mechanism with GAT's attention-based message passing. This model leverages contrastive loss for discriminative feature learning while maintaining attention-based aggregation. The final node representation is computed as:
\begin{equation}
h_v = \lambda h_v^{GAT} + (1 - \lambda) h_v^{CAGN}
\end{equation}
where \( \lambda \) is a tunable weight parameter balancing the two contributions. The fusion mechanism is implemented via a dual-stream network: one branch processes attention-based aggregation, while the other refines embeddings through contrastive loss. The outputs are then adaptively merged using a learnable gating function.

The CAGN module employs multi-head attention-based graph convolutions while incorporating contrastive loss to improve node embeddings by pulling similar nodes closer and pushing dissimilar nodes apart. Given an input node feature matrix \( X \) and edge index \( E \), the first CAGN layer applies an 8-head GAT convolution: $H_1 = \text{ReLU}(\text{GATConv}(X, E))$, where \( H_1 \) represents the hidden embeddings. This is followed by another GAT layer with 4 attention heads for deeper feature extraction: $H_2 = \text{ELU}(\text{GATConv}(H_1, E))$. Finally, a single-head fusion layer aggregates the learned representations: $Z = \text{GATConv}(H_2, E)$, where \( Z \) represents the final output logits. The final hybrid design enhances interpretability and classification performance by leveraging the contrastive loss from CAGN and the structural attention mechanism of GAT.

\begin{table}[!ht]
\begin{center}
\tiny
\begin{minipage}{350pt}
\caption{Performance comparison using adaptive graph construction without augmentation (sorted in terms of macro-average F1 score). \textbf{Bold} indicates the performance of CAGN-GAT Fusion.}
\label{tab:tab1}
\begin{adjustbox}{max width=\textwidth}
\begin{tabular}{clccccccc}
\hline
\multicolumn{1}{l}{\textbf{Dataset}}  & \textbf{Model}             & \textbf{Accuracy} & \textbf{AUC}    & \textbf{Precision} & \textbf{Recall} & \textbf{F1}     & \textbf{Time (s)} & \textbf{Memory (MB)} \\ \hline
\multirow{17}{*}{\textbf{KDD CUP 99\footnotemark[1]}} & CAGN                       & 0.9931            & 0.7987          & 0.9932             & 0.8372          & 0.9019          & 4.76              & 0.19                 \\
                                      & \textbf{\textbf{CAGN-GAT Fusion}} & \textbf{0.9921}   & \textbf{0.7987} & \textbf{0.9929}    & \textbf{0.8361} & \textbf{0.9012} & \textbf{5.54}     & \textbf{0.18}        \\
                                      & GCN                        & 0.9911            & 0.7987          & 0.9912             & 0.7372          & 0.8094          & 2.22              & 0.25                 \\
                                      & SuperGAT                   & 0.9911            & 0.7987          & 0.9912             & 0.7718          & 0.8525          & 127.38            & 5.84                 \\
                                      & GAT                        & 0.9901            & 0.7987          & 0.9625             & 0.7707          & 0.8419          & 3.29              & 0.19                 \\
                                      & MultiScaleGAT              & 0.9891            & 0.7987          & 0.9411             & 0.7207          & 0.7797          & 9.55              & 0.18                 \\
                                      & ClusterGCN                 & 0.9881            & 0.7987          & 0.9401             & 0.7204          & 0.7791          & 1.9               & 0.17                 \\
                                      & GraphSAGE                  & 0.9861            & 0.7987          & 0.9585             & 0.6745          & 0.7491          & 1.28              & 0.19                 \\
                                      & ARMA                       & 0.9842            & 0.7987          & 0.7383             & 0.6091          & 0.6458          & 5.77              & 0.18                 \\
                                      & GIN                        & 0.9347            & 0.7987          & 0.7751             & 0.4831          & 0.5570          & 1.17              & 0.18                 \\
                                      & SVM                        & 0.9327            & 0.8687          & 0.5007             & 0.4314          & 0.4566          & 0.76              & 0.88                 \\
                                      & RF                         & 0.8921            & 0.8054          & 0.6830             & 0.5021          & 0.5419          & 1                 & 0.64                 \\
                                      & XGBoost                    & 0.8901            & 0.8693          & 0.4260             & 0.4030          & 0.4089          & 0.37              & 0.78                 \\
                                      & GB                         & 0.8554            & 0.7987          & 0.5206             & 0.5038          & 0.4682          & 4.67              & 1.23                 \\
                                      & NN                         & 0.7891            & 0.6709          & 0.2907             & 0.3146          & 0.2889          & 4.42              & 1                    \\
                                      & DT                         & 0.3158            & 0.5393          & 0.2794             & 0.2772          & 0.2040          & 0.09              & 0.43                 \\
                                      & LR                         & 0.2812            & 0.2746          & 0.1902             & 0.1307          & 0.1497          & 0.22              & 0.95                 \\ \hline
\multirow{17}{*}{\textbf{UNSW-NB15\footnotemark[2]}}  & GAT                        & 0.9990            & 0.8707          & 0.9722             & 0.9995          & 0.9855          & 2.14              & 0.11                 \\
                                      & MultiScaleGAT              & 0.9980            & 0.8707          & 0.9701             & 0.9701          & 0.9701          & 5.65              & 0.11                 \\
                                      & SuperGAT                   & 0.9980            & 0.8707          & 0.9701             & 0.9701          & 0.9701          & 10.13             & 0.35                 \\
                                      & GIN                        & 0.9980            & 0.8707          & 0.9701             & 0.9701          & 0.9701          & 1.11              & 0.12                 \\
                                      & GraphSAGE                  & 0.9980            & 0.8707          & 0.9701             & 0.9701          & 0.9701          & 1.2               & 0.11                 \\
                                      & GCN                        & 0.9970            & 0.8707          & 0.9677             & 0.9407          & 0.9538          & 1.6               & 0.13                 \\
                                      & ClusterGCN                 & 0.9960            & 0.8707          & 0.9651             & 0.9113          & 0.9365          & 1.6               & 0.11                 \\
                                      & \textbf{\textbf{CAGN-GAT Fusion}} & \textbf{0.9950}   & \textbf{0.8707} & \textbf{0.9623}    & \textbf{0.8818} & \textbf{0.9181} & \textbf{2.22}     & \textbf{0.11}        \\
                                      & ARMA                       & 0.9860            & 0.8707          & 0.7753             & 0.9640          & 0.8442          & 2.59              & 0.11                 \\
                                      & SVM                        & 0.9850            & 0.8418          & 0.8680             & 0.5877          & 0.6391          & 0.57              & 3.94                 \\
                                      & LR                         & 0.9840            & 0.8292          & 0.7930             & 0.5872          & 0.6323          & 0.22              & 2.16                 \\
                                      & RF                         & 0.9830            & 0.8590          & 0.7430             & 0.5867          & 0.6261          & 1.61              & 1.39                 \\
                                      & CAGN                       & 0.9830            & 0.8707          & 0.4915             & 0.5000          & 0.4957          & 3.68              & 0.19                 \\
                                      & NN                         & 0.9810            & 0.7835          & 0.4915             & 0.4990          & 0.4952          & 2.42              & 1.04                 \\
                                      & XGBoost                    & 0.9810            & 0.8165          & 0.6934             & 0.6146          & 0.6433          & 0.38              & 0.19                 \\
                                      & GB                         & 0.9770            & 0.8707          & 0.6179             & 0.5837          & 0.5976          & 7.87              & 1.49                 \\
                                      & DT                         & 0.9660            & 0.6648          & 0.5978             & 0.6648          & 0.6217          & 0.24              & 1.19                 \\ \hline
\multirow{17}{*}{\textbf{CICIDS2017\footnotemark[3]}} & ClusterGCN                 & 0.9850            & 0.8469          & 0.9840             & 0.9221          & 0.9459          & 1.64              & 0.19                 \\
                                      & \textbf{\textbf{CAGN-GAT Fusion}} & \textbf{0.9850}   & \textbf{0.8469} & \textbf{0.9840}    & \textbf{0.9221} & \textbf{0.9459} & \textbf{4.12}     & \textbf{0.19}        \\
                                      & CAGN                       & 0.9850            & 0.8469          & 0.9947             & 0.9223          & 0.9511          & 4.09              & 0.19                 \\
                                      & GAT                        & 0.9840            & 0.8469          & 0.9563             & 0.9218          & 0.9366          & 2.55              & 0.19                 \\
                                      & SuperGAT                   & 0.9820            & 0.8469          & 0.9500             & 0.9007          & 0.9196          & 75.68             & 3.98                 \\
                                      & MultiScaleGAT              & 0.9810            & 0.8469          & 0.9825             & 0.8686          & 0.9017          & 7.36              & 0.19                 \\
                                      & GraphSAGE                  & 0.9741            & 0.8469          & 0.9912             & 0.7599          & 0.8102          & 1.25              & 0.19                 \\
                                      & GCN                        & 0.9721            & 0.8469          & 0.9423             & 0.7748          & 0.8336          & 1.42              & 0.2                  \\
                                      & ARMA                       & 0.9691            & 0.8469          & 0.9902             & 0.7364          & 0.7813          & 4.02              & 0.19                 \\
                                      & GIN                        & 0.9252            & 0.8469          & 0.3136             & 0.3122          & 0.3120          & 1.13              & 0.19                 \\
                                      & RF                         & 0.8194            & 0.8300          & 0.4402             & 0.4345          & 0.4357          & 1.65              & 0.8                  \\
                                      & NN                         & 0.8174            & 0.7961          & 0.2664             & 0.2777          & 0.2718          & 2.99              & 1                    \\
                                      & SVM                        & 0.8094            & 0.8264          & 0.4352             & 0.4293          & 0.4301          & 2.37              & 1.58                 \\
                                      & XGBoost                    & 0.8074            & 0.7775          & 0.3221             & 0.4281          & 0.3390          & 0.7               & 0.88                 \\
                                      & GB                         & 0.7994            & 0.8469          & 0.4906             & 0.4664          & 0.4546          & 18.69             & 1.56                 \\
                                      & DT                         & 0.7535            & 0.6849          & 0.4634             & 0.4591          & 0.4546          & 0.17              & 0.6                  \\
                                      & LR                         & 0.4252            & 0.4355          & 0.0976             & 0.1139          & 0.1047          & 0.23              & 1.35                 \\ \hline
\multirow{17}{*}{\textbf{NSL-KDD\footnotemark[4]}}    & \textbf{\textbf{CAGN-GAT Fusion}} & \textbf{0.9870}   & \textbf{0.8044} & \textbf{0.9795}    & \textbf{0.9879} & \textbf{0.9836} & \textbf{4.4}      & \textbf{0.16}        \\
                                      & ClusterGCN                 & 0.9850            & 0.8044          & 0.9776             & 0.9840          & 0.9807          & 1.62              & 0.16                 \\
                                      & CAGN                       & 0.9810            & 0.8044          & 0.9734             & 0.9770          & 0.9752          & 4.16              & 0.16                 \\
                                      & MultiScaleGAT              & 0.9770            & 0.8044          & 0.9697             & 0.9697          & 0.9697          & 7.63              & 0.17                 \\
                                      & GAT                        & 0.9720            & 0.8044          & 0.9597             & 0.9679          & 0.9637          & 2.63              & 0.16                 \\
                                      & SuperGAT                   & 0.9630            & 0.8044          & 0.9536             & 0.9481          & 0.9507          & 84.08             & 4.01                 \\
                                      & GCN                        & 0.9620            & 0.8044          & 0.9615             & 0.9379          & 0.9482          & 1.38              & 0.17                 \\
                                      & GraphSAGE                  & 0.9500            & 0.8044          & 0.9477             & 0.9193          & 0.9314          & 1.18              & 0.16                 \\
                                      & ARMA                       & 0.9480            & 0.8044          & 0.9477             & 0.9140          & 0.9281          & 4.23              & 0.16                 \\
                                      & GIN                        & 0.8460            & 0.8044          & 0.8791             & 0.7382          & 0.7702          & 1.08              & 0.16                 \\
                                      & SVM                        & 0.7290            & 0.8046          & 0.7358             & 0.6794          & 0.6819          & 2.29              & 0.87                 \\
                                      & GB                         & 0.6490            & 0.8044          & 0.6251             & 0.5679          & 0.5792          & 4.86              & 0.93                 \\
                                      & RF                         & 0.5650            & 0.7686          & 0.4884             & 0.4670          & 0.4657          & 1.52              & 0.63                 \\
                                      & XGBoost                    & 0.5490            & 0.7246          & 0.4566             & 0.4461          & 0.4450          & 0.35              & 0.64                 \\
                                      & DT                         & 0.4750            & 0.5561          & 0.3915             & 0.4007          & 0.3916          & 0.13              & 0.38                 \\
                                      & NN                         & 0.4670            & 0.5482          & 0.3992             & 0.3974          & 0.3900          & 3.72              & 0.94                 \\
                                      & LR                         & 0.2340            & 0.3640          & 0.3055             & 0.2016          & 0.2316          & 0.15              & 0.77                 \\ \hline
\end{tabular}%
\footnotetext[1]{KDD CUP 99 - \url{https://kdd.ics.uci.edu/databases/kddcup99/kddcup99.html}}
\footnotetext[2]{UNSW-NB15 - \url{https://research.unsw.edu.au/projects/unsw-nb15-dataset}}
\footnotetext[3]{CIC-IDS2017 - \url{https://www.unb.ca/cic/datasets/ids-2017.html}}
\footnotetext[4]{NSL KDD - \url{https://www.kaggle.com/datasets/hassan06/nslkdd}}

\end{adjustbox}

\end{minipage}
\end{center}
\end{table}

\begin{table}[!ht]
\begin{center}
\tiny
\begin{minipage}{350pt}
\caption{Performance comparison using adaptive graph construction with augmentation (sorted in terms of macro-average F1 score). \textbf{Bold} indicates the performance of CAGN-GAT Fusion.}
\label{tab:tab2}
\begin{adjustbox}{max width=\textwidth}
\begin{tabular}{llccccccc}
\hline
\textbf{Dataset} & \textbf{Model} & \textbf{Accuracy} & \textbf{AUC} & \textbf{Precision} & \textbf{Recall} & \textbf{F1} & \textbf{Time (s)} & \textbf{Memory (MB)} \\ \hline
\multirow{17}{*}{KDD CUP 99} & \textbf{CAGN-GAT Fusion} & \textbf{0.9871} & \textbf{0.9191} & \textbf{0.9432} & \textbf{0.8246} & \textbf{0.8623} & \textbf{5.96}   & \textbf{0.17} \\
                             & CAGN              & 0.9822 & 0.9191 & 0.9645 & 0.7181 & 0.7867 & 4.87   & 0.18 \\
                             & MultiScaleGAT     & 0.9871 & 0.9191 & 0.9204 & 0.7336 & 0.7796 & 9.88   & 0.17 \\
                             & ClusterGCN        & 0.9832 & 0.9191 & 0.9372 & 0.7183 & 0.7766 & 1.83   & 0.17 \\
                             & GCN               & 0.9762 & 0.9191 & 0.9579 & 0.6127 & 0.6855 & 1.6    & 0.18 \\
                             & SuperGAT          & 0.9802 & 0.9191 & 0.8344 & 0.6348 & 0.6853 & 129.29 & 5.98 \\
                             & GraphSAGE         & 0.9782 & 0.9191 & 0.9806 & 0.5923 & 0.6804 & 1.27   & 0.17 \\
                             & GAT               & 0.9822 & 0.9191 & 0.9313 & 0.6159 & 0.6780 & 3.4    & 0.17 \\
                             & ARMA              & 0.9772 & 0.9191 & 0.7790 & 0.5415 & 0.5990 & 6.01   & 0.17 \\
                             & RF                & 0.9069 & 0.8240 & 0.5529 & 0.4931 & 0.5072 & 0.97   & 0.63 \\
                             & GB                & 0.9050 & 0.9191 & 0.5254 & 0.4713 & 0.4825 & 4.64   & 1.21 \\
                             & SVM               & 0.9356 & 0.8880 & 0.5056 & 0.4348 & 0.4607 & 0.8    & 0.88 \\
                             & DT                & 0.8089 & 0.7028 & 0.4379 & 0.4697 & 0.4445 & 0.09   & 0.43 \\
                             & NN                & 0.9347 & 0.9573 & 0.3984 & 0.4092 & 0.4008 & 3.34   & 0.97 \\
                             & XGBoost           & 0.8812 & 0.9613 & 0.3520 & 0.4075 & 0.3723 & 0.35   & 0.72 \\
                             & LR                & 0.9277 & 0.6655 & 0.3711 & 0.3542 & 0.3615 & 0.18   & 0.91 \\
                             & GIN               & 0.7267 & 0.9191 & 0.2467 & 0.2732 & 0.2542 & 1.22   & 0.17 \\ \hline
\multirow{17}{*}{UNSW-NB15}  & GIN               & 0.9980 & 0.8809 & 0.9701 & 0.9701 & 0.9701 & 1.1    & 0.11 \\
                             & SuperGAT          & 0.9970 & 0.8809 & 0.9677 & 0.9407 & 0.9538 & 10.5   & 0.34 \\
                             & GAT               & 0.9970 & 0.8809 & 0.9677 & 0.9407 & 0.9538 & 2.12   & 0.11 \\
                             & MultiScaleGAT     & 0.9960 & 0.8809 & 0.9651 & 0.9113 & 0.9365 & 5.57   & 0.11 \\
                             & GraphSAGE         & 0.9960 & 0.8809 & 0.9651 & 0.9113 & 0.9365 & 1.19   & 0.11 \\
                             & \textbf{CAGN-GAT Fusion} & \textbf{0.9950} & \textbf{0.8809} & \textbf{0.9623} & \textbf{0.8818} & \textbf{0.9181} & \textbf{2.21}   & \textbf{0.11} \\
                             & GCN               & 0.9950 & 0.8809 & 0.9623 & 0.8818 & 0.9181 & 1.38   & 0.12 \\
                             & ClusterGCN        & 0.9920 & 0.8809 & 0.9510 & 0.7936 & 0.8551 & 1.57   & 0.11 \\
                             & DT                & 0.9750 & 0.6983 & 0.6540 & 0.6983 & 0.6731 & 0.23   & 1.19 \\
                             & ARMA              & 0.9360 & 0.8809 & 0.6007 & 0.9385 & 0.6499 & 2.5    & 0.11 \\
                             & SVM               & 0.9850 & 0.8149 & 0.8680 & 0.5877 & 0.6391 & 0.54   & 3.94 \\
                             & RF                & 0.9820 & 0.8787 & 0.7072 & 0.5862 & 0.6204 & 1.58   & 1.39 \\
                             & XGBoost           & 0.9750 & 0.8199 & 0.6184 & 0.6115 & 0.6149 & 0.37   & 0.2  \\
                             & GB                & 0.9790 & 0.8809 & 0.6429 & 0.5847 & 0.6058 & 8.09   & 1.49 \\
                             & LR                & 0.9810 & 0.5512 & 0.6591 & 0.5568 & 0.5822 & 0.22   & 2.15 \\
                             & CAGN              & 0.9830 & 0.8809 & 0.4915 & 0.5000 & 0.4957 & 3.58   & 0.18 \\
                             & NN                & 0.9830 & 0.4458 & 0.4915 & 0.5000 & 0.4957 & 1.54   & 1.04 \\ \hline
\multirow{17}{*}{CICIDS2017} & \textbf{CAGN-GAT Fusion} & \textbf{0.9751} & \textbf{0.8431} & \textbf{0.9823} & \textbf{0.8554} & \textbf{0.8812} & \textbf{4.32}   & \textbf{0.19} \\
                             & MultiScaleGAT     & 0.9741 & 0.8431 & 0.9815 & 0.8441 & 0.8749 & 7.57   & 0.19 \\
                             & CAGN              & 0.9741 & 0.8431 & 0.9820 & 0.8346 & 0.8515 & 4.22   & 0.19 \\
                             & GAT               & 0.9721 & 0.8431 & 0.9259 & 0.7715 & 0.8190 & 2.61   & 0.19 \\
                             & SuperGAT          & 0.9711 & 0.8431 & 0.7935 & 0.8024 & 0.7969 & 80.31  & 3.99 \\
                             & GCN               & 0.9671 & 0.8431 & 0.9904 & 0.7462 & 0.7900 & 1.41   & 0.19 \\
                             & GraphSAGE         & 0.9571 & 0.8431 & 0.8210 & 0.5746 & 0.6247 & 1.21   & 0.19 \\
                             & ClusterGCN        & 0.9581 & 0.8431 & 0.7171 & 0.6037 & 0.5954 & 1.63   & 0.19 \\
                             & ARMA              & 0.9481 & 0.8431 & 0.6522 & 0.5297 & 0.5552 & 4.21   & 0.19 \\
                             & GB                & 0.7934 & 0.8431 & 0.4621 & 0.4653 & 0.4593 & 18.52  & 1.54 \\
                             & RF                & 0.8104 & 0.8822 & 0.4782 & 0.4508 & 0.4590 & 1.59   & 0.81 \\
                             & SVM               & 0.8094 & 0.8199 & 0.4349 & 0.4293 & 0.4300 & 2.16   & 1.58 \\
                             & NN                & 0.7944 & 0.8100 & 0.4263 & 0.4258 & 0.4250 & 3.56   & 1    \\
                             & DT                & 0.7226 & 0.6647 & 0.3879 & 0.4335 & 0.3953 & 0.16   & 0.6  \\
                             & XGBoost           & 0.7874 & 0.8556 & 0.3102 & 0.4233 & 0.3323 & 0.69   & 0.88 \\
                             & GIN               & 0.8094 & 0.8431 & 0.2910 & 0.2542 & 0.2591 & 1.11   & 0.19 \\
                             & LR                & 0.5170 & 0.3044 & 0.1331 & 0.1449 & 0.1356 & 0.2    & 1.35 \\ \hline
\multirow{17}{*}{NSL-KDD}    & GraphSAGE         & 0.9060 & 0.8368 & 0.8837 & 0.8628 & 0.8720 & 1.21   & 0.16 \\
                             & ARMA              & 0.9070 & 0.8368 & 0.8902 & 0.8578 & 0.8710 & 4.43   & 0.16 \\
                             & ClusterGCN        & 0.9030 & 0.8368 & 0.8865 & 0.8488 & 0.8641 & 1.61   & 0.16 \\
                             & \textbf{CAGN-GAT Fusion} & \textbf{0.8970} & \textbf{0.8368} & \textbf{0.8731} & \textbf{0.8528} & \textbf{0.8620} & \textbf{4.62}   & \textbf{0.16} \\
                             & GAT               & 0.8930 & 0.8368 & 0.8711 & 0.8345 & 0.8483 & 2.75   & 0.16 \\
                             & CAGN              & 0.8870 & 0.8368 & 0.8634 & 0.8302 & 0.8432 & 4.29   & 0.16 \\
                             & SuperGAT          & 0.8480 & 0.8368 & 0.8028 & 0.7696 & 0.7832 & 91.85  & 4.13 \\
                             & GCN               & 0.8130 & 0.8368 & 0.7733 & 0.7470 & 0.7533 & 1.41   & 0.17 \\
                             & MultiScaleGAT     & 0.8010 & 0.8368 & 0.7708 & 0.7489 & 0.7494 & 7.96   & 0.16 \\
                             & GB                & 0.7620 & 0.8368 & 0.7293 & 0.7013 & 0.7117 & 4.89   & 0.93 \\
                             & SVM               & 0.7480 & 0.8197 & 0.7472 & 0.6908 & 0.6964 & 2.32   & 0.87 \\
                             & XGBoost           & 0.6520 & 0.7992 & 0.6139 & 0.5771 & 0.5862 & 0.35   & 0.61 \\
                             & RF                & 0.6520 & 0.8164 & 0.6342 & 0.5727 & 0.5859 & 1.55   & 0.63 \\
                             & GIN               & 0.6110 & 0.8368 & 0.6329 & 0.5472 & 0.5708 & 1.13   & 0.16 \\
                             & DT                & 0.6180 & 0.6665 & 0.5914 & 0.5547 & 0.5626 & 0.11   & 0.38 \\
                             & NN                & 0.6080 & 0.7516 & 0.5825 & 0.5278 & 0.5290 & 2.95   & 0.94 \\
                             & LR                & 0.5440 & 0.5247 & 0.4482 & 0.4157 & 0.4110 & 0.15   & 0.77 \\ \hline
\end{tabular}%
\end{adjustbox}
\end{minipage}
\end{center}
\end{table}

\subsection{Experimental Setup}

\subsubsection{Datasets Used}:
We use four benchmark intrusion detection datasets; among them, only UNSW-NB15 was used for binary classification, and the others were used for multiclass classification.

\begin{enumerate}
\item \textbf{NSL-KDD:} This dataset is an improved version of the KDD Cup 1999 dataset for network intrusion detection. It contains 41 features, including duration, protocol type, service, bytes, and flags, with the target variable being the attack type. The dataset is used to classify network traffic into normal and attack types. Preprocessing includes feature selection and balancing class distribution by grouping less frequent attack types into one class.
\item \textbf{UNSW-NB15:} This dataset has 49 features, including packet-level statistics, flow characteristics, and network connection details, with the target being the attack label. It includes different attack types such as DoS, Probe, and Exploit. The preprocessing steps include removing high-correlation features and creating new features like network bytes for better model performance.
\item \textbf{CICIDS2017:} This dataset is collected from various network traffic scenarios, including different types of attacks such as DoS, DDoS, and infiltration attempts. It provides several features related to flow data and connection statistics. The target variable is also the attack type. Preprocessing steps include handling missing values, feature scaling, and converting categorical features into numerical representations for model compatibility.
\item \textbf{KDD Cup 1999:} This classic dataset contains network traffic data labeled as either normal or one of several attack types, like neptune, smurf, and back. It comprises 41 features, including protocol type, service, and number of failed logins. Preprocessing involves extracting useful features and dealing with imbalanced class distribution by reclassifying some attacks as a single class.
\end{enumerate}

\subsubsection{Implementation Details}:
To ensure data consistency and integrity, our research involves preprocessing multiple cybersecurity datasets, including NSL-KDD, UNSW-NB15, CICIDS2017, and KDD CUP 99. We handle missing values by dropping rows with NaNs in numeric and categorical columns, encoding categorical features using Label Encoding, and normalizing numerical features with StandardScaler. To maintain class imbalance while reducing dataset size, we employ a proportional downsampling approach where large classes are scaled while small classes remain unchanged. We introduce feature correlation by applying a randomized transformation matrix with a correlation level of 0.9 and weaken feature predictability by retaining only the least informative 30\% based on mutual information. 

The data is split into an 80:20 train-test ratio, followed by adaptive graph construction using an Euclidean-based metric with a threshold of 0.5. Additionally, data augmentation techniques, including 10\% edge perturbation and 20\% feature masking, are applied to enhance the robustness of the constructed graphs. Finally, the processed datasets and graphs are utilized to benchmark various GNN models. We selected baselines, including Logistic Regression (LR), DT, Multilayer Perceptron-based Neural Network (NN), SVM, RF, XGBoost, and Gradient Boosting (GB), to benchmark the performance of the GNN models across linear models, tree-based methods, and DL. These models demonstrated state-of-the-art performance in previous studies on the tabular datasets used. However, we aimed to investigate their performance on short and imbalanced datasets. 

The training process involves optimizing the model using the Adam optimizer with an initial learning rate of 0.001, while a `CosineAnnealingLR' scheduler gradually adjusts the learning rate over 200 cycles. GNN models are trained for 300 epochs using binary cross-entropy or cross-entropy loss, depending on the classification task. After each epoch, gradients are backpropagated, and parameters are updated. The model's predictions are processed during evaluation using sigmoid activation for binary classification and softmax for multi-class cases. The code is publicly available at \href{https://github.com/Abrar2652/Network-Intrusion-Detection}{https://github.com/Abrar2652/Network-Intrusion-Detection}.

\section{Computational Results and Discussion}
\label{sec:resultdiscussion}
The main objective of this study was to benchmark state-of-the-art GNNs and traditional ML models for network intrusion detection and to analyze the impact of adaptive graph construction (see Table~\ref{tab:tab1}) and graph augmentation (see Table~\ref{tab:tab2}) strategies on model performance and efficiency. The proposed CAGN-GAT Fusion model demonstrated consistently competitive performance across four benchmark datasets. Without augmentation, it achieved top-tier results on KDD CUP 99 (accuracy: 0.9921, F1: 0.9012), NSL-KDD (accuracy: 0.9870, F1: 0.9836), and tied for the highest score on CICIDS2017 (accuracy: 0.9850, F1: 0.9459). However, on UNSW-NB15, its F1-score (0.9181) lagged behind GAT (0.9855), likely due to the dataset’s sparse graph structure and severe class imbalance, which favor simpler attention aggregation mechanisms over contrastive-based fusion. When applying graph augmentation (10\% edge perturbation and 20\% feature masking), CAGN-GAT Fusion maintained strong performance on KDD CUP 99 (accuracy: 0.9871, F1: 0.8623) and CICIDS2017 (F1: 0.8812) but experienced a noticeable decline on NSL-KDD (F1: 0.8620), attributed to noisy synthetic connections disrupting meaningful graph neighborhoods. In terms of resource efficiency, the model consistently exhibited low memory usage (0.11–0.19 MB) while achieving better precision-recall trade-offs than faster, lightweight models like ClusterGCN and avoiding the high computational overhead of models such as SuperGAT (up to 5.98 MB). Key insights reveal that contrastive learning benefits from dense, balanced graphs, while vanilla GAT performs better on sparse, imbalanced data; additionally, augmentation improves generalization in balanced datasets but may introduce harmful noise in skewed ones. These findings highlight the importance of dataset characteristics in selecting graph learning strategies and suggest future directions, including adaptive augmentation rates, dynamic attention mechanisms, and integrating GraphSAGE to improve performance on sparse and imbalanced intrusion detection data.

\section{Conclusions and Future Works}
\label{sec:conclusion}
This study explored different GNN models for network intrusion detection and proposed CAGN-GAT Fusion as the best and most generalizable performer across all datasets. It achieved competitive and robust accuracy, precision, recall, and F1 score, proving its effectiveness in detecting cyber threats with less computational time and memory requirements. GNNs demonstrated superior ability in learning complex network attack patterns compared to traditional ML models. Our research also showed that graph augmentation further improves performance, particularly in handling imbalanced datasets. However, we noticed that some models require high computational resources, which may not be ideal for real-time applications. Future work can further focus on trying advanced feature selection methods before graph construction. Moreover, multiview graph construction techniques can be explored where multiple nearest neighbors and distance metrics can be considered in graph data. CAGN-GAT Fusion can be tuned further using more heads in the GCNConv layers and evaluating the effect of adding a GraphSage layer, since GraphSage showed consistency in all cases. We also aim to improve model adaptability to new and evolving cyber threats by incorporating dynamic graph structures and self-learning techniques. Further studies can also investigate how GNNs can be deployed in real-world security systems with minimal latency and resource consumption.


{
\small
\bibliographystyle{splncs04}
\bibliography{main}
}

\end{document}